\documentclass[runningheads]{llncs}

\usepackage[T1]{fontenc}
\usepackage{graphicx}
\usepackage{orcidlink}
\usepackage{amsmath}
\usepackage{adjustbox}
\usepackage{multirow}
\usepackage{array}
\usepackage{pifont}
\usepackage{float}
\usepackage{booktabs,tabularx}

\newcommand{\scriptD}{\mathcal{D}}

\usepackage{color}

\begin{document}
\title{Multi-LLM Collaborative Caption Generation in Scientific Documents}
%
%
\author{Jaeyoung Kim\inst{1}\orcidlink{0000-0003-0880-0398} \and
Jongho Lee\inst{1}\orcidlink{0000-0002-8520-2722} \and
Hong-Jun Choi\inst{1}\orcidlink{0000-0002-3413-511X} \and
Ting-Yao Hsu\inst{2}\orcidlink{0009-0008-9082-6039} \and
Chieh-Yang Huang\inst{2}\orcidlink{0009-0001-6736-9959} \and
Sungchul Kim\inst{3}\orcidlink{0000-0003-3580-5290} \and
Ryan Rossi\inst{3} \and
Tong Yu\inst{3}\orcidlink{0000-0002-5991-2050} \and
Clyde Lee Giles\inst{2}\orcidlink{0000-0002-1931-585X} \and
Ting-Hao `Kenneth' Huang\inst{2}\orcidlink{0000-0001-7021-4627} \and
Sungchul Choi\inst{1}\orcidlink{0000-0002-5836-3838}}
\authorrunning{J. Kim et al.}
%
\institute{Teamreboott Inc., Busan, Korea \\
\email{\{jaeyoungkim,jongho.lee,hongjun.choi,admin\}@reboott.ai} \\ \and
Pennsylvania State University, University Park, PA, USA. \\
\email{\{txh357,chiehyang,clg20,txh710\}@psu.edu} \\ \and
Adobe Research, San Francisco, CA, USA. \\ 
\email{\{sukim,ryrossi,tyu\}@adobe.com}}
\maketitle              
\begin{abstract}
Scientific figure captioning is a complex task that requires generating contextually appropriate descriptions of visual content. However, existing methods often fall short by utilizing incomplete information, treating the task solely as either an image-to-text or text summarization problem. This limitation hinders the generation of high-quality captions that fully capture the necessary details. Moreover, existing data sourced from arXiv papers contain low-quality captions, posing significant challenges for training large language models (LLMs).
In this paper, we introduce a framework called \textbf{M}ulti-\textbf{L}LM Colla\textbf{b}orative Figure \textbf{Cap}tion Generation (\textbf{MLBCAP)} to address these challenges by leveraging specialized LLMs for distinct sub-tasks. Our approach unfolds in three key modules: (\textbf{Quality Assessment}) We utilize multimodal LLMs to assess the quality of training data, enabling the filtration of low-quality captions. (\textbf{Diverse Caption Generation}) We then employ a strategy of fine-tuning/prompting multiple LLMs on the captioning task to generate candidate captions.  (\textbf{Judgment}) Lastly, we prompt a prominent LLM to select the highest quality caption from the candidates, followed by refining any remaining inaccuracies. Human evaluations demonstrate that informative captions produced by our approach rank better than human-written captions, highlighting its effectiveness. Our code is available at~\url{https://github.com/teamreboott/MLBCAP}

\end{abstract}
\section{Introduction}

Scientific figures are integral to academic communication, offering a concise and effective means of presenting complex information. However, the value of a figure is largely determined by the quality of its accompanying caption. Captions provide essential context, elucidate visual elements, and enable readers to fully grasp the insights conveyed by the figure. Consequently, the generation of accurate and informative captions for scientific documents is critical to effectively communicating key findings to domain experts. Automated captioning not only aids researchers by improving the clarity of figure descriptions but also contributes to the overall enhancement of scholarly communication~\cite{hsu2024scicapenter}.

Existing approaches to automatic figure captioning have predominantly treated the task either as an image-to-text problem~\cite{SciCap,qian2021generating} or a text summarization task~\cite{SciCapSum,chao2024solution}. Image-to-text methods focus on extracting information directly from visual content, but they often lack the domain-specific understanding required to interpret abbreviations, symbols, and implicit relationships. On the other hand, text summarization approaches rely on textual metadata such as figure-mentioning paragraphs or optical character recognition (OCR) outputs from figures. While these methods can capture textual context, they frequently overlook crucial visual details, such as trends, patterns, and color-coded elements that are vital for a comprehensive understanding of the figure. Consequently, these fragmented approaches fail to produce captions that are both accurate and informative, underscoring the need for a unified framework capable of leveraging both textual and visual modalities.

Another major challenge in figure captioning lies in the quality of available training data. Many existing datasets~\cite{kahou2017figureqa,kafle2018dvqa,SciCap,yang2023scicap+}, particularly those sourced from platforms like arXiv, contain captions that are incomplete, verbose, or poorly written. 
A recent study~\cite{SciCapSum} reports that over 50\% of captions in arXiv papers are unhelpful to domain experts. These low-quality captions can hinder model training and result in sub-optimal caption generation, complicating the accurate assessment of model performance.

To this end, we propose a unified framework named \textbf{M}ulti-\textbf{L}LM Colla\textbf{b}orative Figure \textbf{Cap}tion Generation (\textbf{MLBCAP}). Unlike previous methods, MLBCAP integrates textual and visual information through a carefully orchestrated pipeline comprising three key components: quality assessment, diverse caption generation, and judgment. The quality assessment module filters out low-quality training captions, ensuring that the models are trained on reliable data. In the caption generation stage, multiple LLMs, each specializing in different aspects of figure captioning, collaborate to produce diverse candidate captions. Finally, a judgment module utilizes a prominent LLM to select the best candidate caption and refine it for accuracy and coherence.

While prior studies find that longer captions are generally more beneficial to readers~\cite{hartley2003single,SciCapSum}, scientific journals and conference papers often impose strict page limits. To accommodate this, our framework is designed to generate both long and short versions of captions. In the final step, we utilize GPT-4o with specific instructions regarding caption length to achieve that both versions are concise yet informative. In a human evaluation by domain experts, captions generated by our method are preferred over the original author-written captions, demonstrating the effectiveness of our approach. Our main contributions are as follows:

\begin{itemize}
    \item We propose a unified framework that includes data cleaning, caption generation, and post-editing processes to generate high-quality captions.
    \item Our approach integrates both textual and visual features, leveraging multi-modal models to produce contextually rich and accurate captions.
    \item Through human evaluations, we show that our approach ranked better than author-written captions, demonstrating its effectiveness. 
\end{itemize}

\section{Related Work}

\subsection{Collaboration Techniques with LLMs} 

LLMs have shown exceptional performance across a wide range of tasks, benefiting from their ability to comprehend instructions~\cite{GPT4,YI}. However, despite their versatility, individual LLMs exhibit distinct strengths and limitations due to differences in training data and architectural design~\cite{jiang-etal-2023-llmlingua}. To mitigate this issue, recent work~\cite{si2023getting} trained a classifier to select the best response generated by different reasoning models. Another related work~\cite{LLM_ENSEMBLE} proposed an algorithm that combines outputs from multiple LLMs for attribute extraction through weight assignment. Despite the growing application of collaborative methods in various fields, a significant research gap exists in exploring their potential for figure captioning in scientific documents.

\subsection{Figure Captioning in Scientific Documents}

To facilitate the generation of captions by neural networks, previous research developed a variety of datasets, such as FigureSeer~\cite{siegel2016figureseer}, FigureQA~\cite{kahou2017figureqa}, DVQA~\cite{kafle2018dvqa}, and SciCap~\cite{SciCap}. More recently, an enhanced version of SciCap was introduced, incorporating both figures and their associated textual information~\cite{SciCapSum}. This dataset advances caption generator capabilities, enabling them to produce contextually relevant captions for scientific figures. Based on this dataset, a recent study~\cite{SciCapSum} discovered that more than 76\% of the words in figure captions matched those in figure-mentioning paragraphs and OCR text. Based on this empirical observation, they formulated the figure captioning task as a text summarization task. Contemporaneously, SciCap+~\cite{yang2023scicap+} was proposed as an extension of SciCap, integrating OCR-derived textual data to further enhance the generation of figure captions. However, text summarization models, which depend on textual data from figure-related paragraphs and OCR outputs, often fail to capture essential visual details, including patterns and colors in graphs.

\subsection{Evaluating Natural Language Generation (NLG) Tasks} 

In NLG tasks, traditional automatic metrics such as BLEU~\cite{BLEU} and ROUGE~\cite{ROUGE} are widely used for evaluation. However, these metrics often exhibit a relatively low correlation with human judgments in text generation tasks~\cite{NLG_EVAL}, mainly because they depend on human-preferred reference outputs to fairly evaluate the performance of NLG models. Recent studies have advocated for using LLMs as reference-free evaluation metrics, achieving higher correspondence with human evaluations than traditional metrics~\cite{UniEval,NLG_EVAL}. Notably, SciCap-Eval~\cite{SciCapEval} employed LLMs to assess caption quality and demonstrated that GPT-4, as a zero-shot caption evaluator, positively correlates with Ph.D. students' assessments (Pearson correlation coefficient of 0.5).

\begin{figure}[t]
\centering
\includegraphics[width=\textwidth]{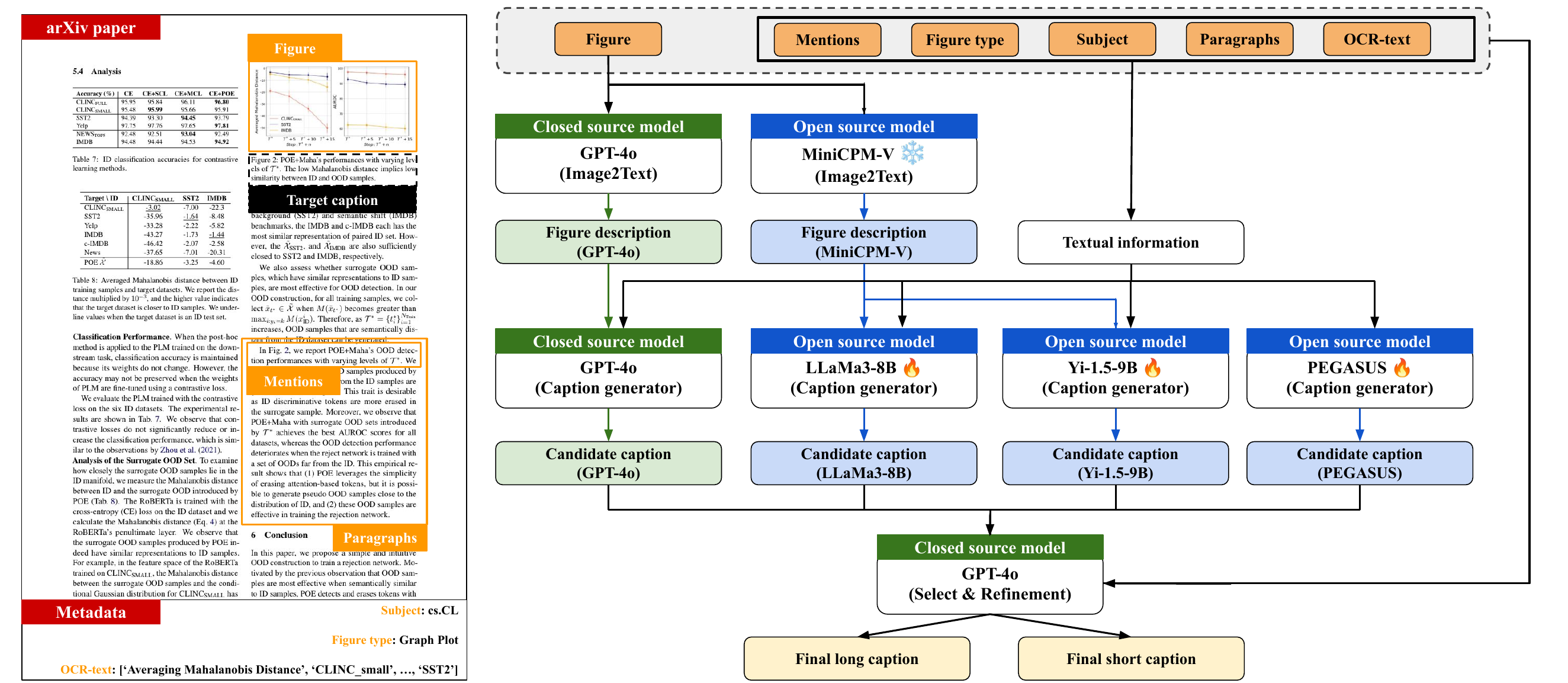}

\caption{Overview of the collaborative framework integrating multiple LLMs for caption generation in scientific documents. Initially, two MLLMs generate figure descriptions. Next, three fine-tuned models and GPT-4o generate candidate captions. Finally, GPT-4o selects and refines the best caption from the candidates.}

\label{fig:overall_framework}
\end{figure}

\section{Problem Statement}
\label{sec:problem}

Consider an arXiv paper $D$ with $n$ captions $\{C_i\}_{i=1}^n$. For each caption $C_i$, several related sources from $D$ are used to assist in caption generation. These sources include the corresponding figure $F_i$ and $m$ paragraphs $\{P_i^j\}_{j=1}^m$ that mention $F_i$. Specifically, $k$ sentences $\{M_i^j\}_{j=1}^k$ within these paragraphs explicitly refer to $F_i$ (e.g., ``As shown in Fig. 1, ...'').~\footnote{For simplicity, superscripts for $P_i$ and $M_i$ are omitted in the following sections, as multiple instances may exist.} Additionally, textual information extracted using OCR from the figure is denoted as $O_i$, and the figure’s type (e.g., bar chart, node diagram) is represented as $T_i$. Finally, the subject category of $D$ (e.g., ``cs.AI'' for Computer Science - Artificial Intelligence) is denoted as $S$. 
The objective of this work is to generate high-quality $C_i$ for $F_i$ utilizing the aforementioned figure-relevant features.

\section{Multi-LLM Collaborative Figure Caption Generation}
\label{sec:method}
Our overall pipeline is illustrated in Figure~\ref{fig:overall_framework}, and the following sections provide a description of each component. The actual prompts are described in the Appendix.

\subsection{Quality Assessment}

We employ GPT-4o to generate a synthetic quality assessment dataset using 3k subset of the training data. Following the approach of SciCap-Eval, we prompt GPT-4o to score captions on a scale of 1 to 6 based on the given $C_i$, $F_i$, and $P_i$ (higher scores indicate better quality). 

Next, we fine-tune LLaVA~\cite{LLaVA} on the constructed dataset. After fine-tuning, LLaVA predicts the caption quality across the entire training dataset. We then collect samples $\scriptD_{\text{high}}$ with quality scores of 5 and 6.
For the evaluation of 200 samples, the fine-tuned LLaVA showed agreement in quality assessment with GPT-4o, achieving Kendall's tau coefficient of 0.5502.

\subsection{Diverse Caption Generation}

To capture diverse perspectives and generate a varied set of candidate captions, we utilize four distinct models: GPT-4o, LLaMA-3-8B~\cite{LLAMA3}, Yi-1.5-9B~\cite{YI}, and Pegasus~\cite{PEGASUS}. Each model offers unique viewpoints that contribute to the diversity of the generated captions. Specifically, GPT-4o leverages its advanced reasoning capabilities as a large-scale LLM. LLaMA-3-8B and Yi-1.5-9B are fine-tuned on the figure captioning task, enhancing domain-specific knowledge. Pegasus excels in abstractive summarization of textual content, capturing essential information from figure-mentioning paragraphs and OCR text.

\noindent \textbf{\textit{GPT-4o}}. For a test sample, we use few-shot prompting by providing GPT-4o with ten example captions, $E$, randomly selected from $\scriptD_{high}$. These examples have the same subject as the test sample and have a quality score of 6. We first instruct GPT-4o to generate a figure description $Z_i$ for $F_i$ by providing the $F_i$, $T_i$ and $S$. Then, GPT-4o generates a candidate caption based on $E$, $P_i$, $M_i$, $T_i$, $O_i$, $S$, and $Z_i$.

\noindent \textbf{\textit{LLaMA-3-8B and Yi-1.5-9B}} are fine-tuned with visual and textual features from the $\scriptD_{\text{high}}$ dataset. Figure descriptions are generated by MiniCPM-V~\cite{MINICPMV}, which outperforms GPT-4V-1106 and Gemini-Pro~\cite{GEMINI} for OpenCompass benchmarks~\cite{OPENCOMPASS}. The prompts used for LLaMA-3-8B and Yi-1.5-9B are identical to the prompts used for GPT-4o, except for the exclusion of the few-shot examples.

\noindent \textbf{\textit{Pegasus}} is fine-tuned on figure-mentioning paragraphs and OCR-text from $\scriptD_{\text{high}}$. Apart from the dataset, this model follows the previous work~\cite{SciCapSum}.

\subsection{Judgement}

We ask GPT-4o to select the best quality caption from four candidate captions and edit inaccuracies in the selected caption leveraging both visual and textual information.
To generate both long and short captions, we set a word limit using the placeholder \verb|[MAX_LEN]| in the prompt. 
To maintain the conciseness while sufficiently conveying figure information, we define \verb|[MAX_LEN]| as 50 words for long captions and 30 words for short captions. These limits are based on the training data, where the average caption length is 41.85 words.
Here, we refer to the generation of long captions as MLBCAP (long) and short captions as MLBCAP (short).

\begin{table}[t]
\centering
\caption{Statistics of the original and preprocessed datasets used for training, validation, and testing.}
\begin{adjustbox}{width=6.0cm,center}
\renewcommand{\arraystretch}{1.3}
\begin{tabular}{lccc}
\hline
& \textbf{Train} & \textbf{Validation} & \textbf{Test} \\
\hline
SciCap   & 360,340 & 47,639 & 47,639 \\
SciCap+ & 394,005 & -      & -      \\
\hline
Preprocessed & 135,935 & 47,639      & 47,639 \\
\hline
\end{tabular}
\end{adjustbox}
\label{tab:dataset_statistic}
\end{table}

\begin{table}[t]
\centering
\caption{The result of caption quality evaluation using GPT-4o. The highest quality captions have a low percentage of 27.11\%.}
\begin{adjustbox}{width=7.0cm,center}
\renewcommand{\arraystretch}{1.3}
\begin{tabular}{lcccccc} 
\hline
\textbf{Score} & \textbf{1} & \textbf{2} & \textbf{3} & \textbf{4} & \textbf{5} & \textbf{6} \\ \hline
N (\%) & \begin{tabular}[c]{@{}c@{}}157\\ (5.24)\end{tabular} & \begin{tabular}[c]{@{}c@{}}305\\ (10.13)\end{tabular} & \begin{tabular}[c]{@{}c@{}}102\\ (3.38)\end{tabular} & \begin{tabular}[c]{@{}c@{}}166\\ (5.53)\end{tabular} & \begin{tabular}[c]{@{}c@{}}1,457\\ (48.58)\end{tabular} & \begin{tabular}[c]{@{}c@{}}813\\ (27.11)\end{tabular} \\ \hline
\end{tabular}
\end{adjustbox}
\label{tab:quality_dataset}
\end{table}

\section{Experiments}

\subsection{Dataset}
One of the goals of this study is to evaluate whether MLBCAP can generate captions that align with human preferences. To enhance the richness of the training data, we have chosen to combine the SciCap\footnote{\url{https://huggingface.co/datasets/CrowdAILab/scicap/tree/main}} and SciCap+~\cite{yang2023scicap+} datasets for training.
These datasets include caption-relevant metadata extracted from the original arXiv papers. 
Detailed statistics of the datasets is provided in Table~\ref{tab:dataset_statistic}.

\noindent \textbf{Data Preprocessing.} Since both SciCap and SciCap+ are built on arXiv papers, we first conducted deduplication using the arXiv paper ID. To obtain high-quality examples, the following preprocessing steps were applied. First, figures with captions that do not end with a period were excluded. Second, captions that were either too long or too short, specifically those exceeding 100 words in length or consisting of a single sentence, were removed. Lastly, figures with captions rated below 5, as described in Section \ref{sec:method}, were filtered out. After these preprocessing steps, we could get 135,935 samples. 


\noindent \textbf{Caption Quality Dataset.} We employ GPT-4o to generate a synthetic quality assessment dataset using 3k subset samples for the training data since there is no publicly available dataset.
The quality distribution of the 3k captions is shown in Table~\ref{tab:quality_dataset}, which shows that the training data contains low-quality captions.

\begin{table}[t]
\centering
\caption{The example of figure descriptions between GPT-4o and MiniCPM-V.}
\begin{adjustbox}{width=\textwidth,center}
\renewcommand{\arraystretch}{1.3}
\begin{tabular}{>{\centering\arraybackslash}m{3cm} |>{\arraybackslash}m{5cm} |>{\arraybackslash}m{5cm}} \hline
\textbf{Figure} & \centering\arraybackslash\textbf{GPT-4o} & \centering\arraybackslash\textbf{MiniCPM-V} \\ \hline 
\centering \includegraphics[width=3cm]{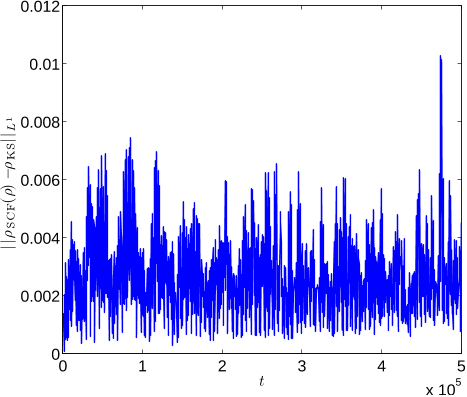}
 & The graph shows the L1 norm of the difference between $pSCF(t)$ and $pKs$ over time (t). The x-axis represents time (t) ranging from 0 to 500,000, and the y-axis represents the L1 norm of the difference, ranging from 0 to 0.012. The plot indicates fluctuations in the difference over time, with occasional peaks.
 & The image contains a graph with a blue line representing the plot of data points. The x-axis is labeled as x $10^5$ and the y-axis is labeled as $||pSCF(p) - pKs||$. There are no additional textual elements or objects present in the image. \\ \hline

 \centering \includegraphics[width=3cm]{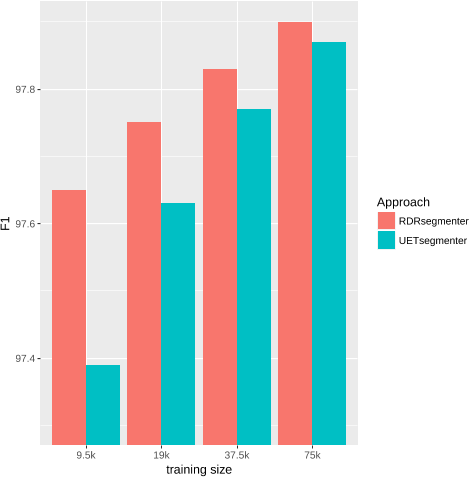}
 & The bar chart compares the F1 scores of two segmentation approaches, RDRsegmenter and UETsegmenter, across different training sizes (9.5k, 19k, 37.5k, and 75k). RDRsegmenter consistently achieves higher F1 scores than UETsegmenter at all training sizes. Both approaches show an increase in F1 score as the training size increases, with RDRsegmenter reaching the highest F1 score at 75k training size.
 & The image contains a bar chart with two types of bars representing different approaches: RDRsegmenter and UETsegmenter. The x-axis represents the training size (k), and the y-axis represents the F1 score, which ranges from 97.4 to 98. \\ \hline
 
\end{tabular}
\end{adjustbox}
\label{tab:description}
\end{table}

\subsection{Training Details}
\label{sec:appendix_train}

\noindent \textbf{Prompts}. The specific prompts employed in our study are detailed in the Appendix. To elicit figure descriptions from the MiniCPM-V model, we employed a direct and intuitive prompt: ``\textit{What is in the image?}''. We also tried extracting figure descriptions with the prompts used in GPT-4o, however, the quality of the generated descriptions was inferior compared to descriptions obtained using the more intuitive prompts. A representative example of the figure descriptions generated is illustrated in Table~\ref{tab:description}.

\noindent \textbf{Models}. Here are the models we used in our experiments. The fine-tuned models were tuned for 5 epochs.

\begin{itemize}
    \item \textbf{GPT-4o.} We used \texttt{gpt-4o-2024-05-13} with a temperature setting of 0. All experimental procedures were conducted between May 2024 and August 2024.
    \item \textbf{LLaVA.} We used \verb|llava-llama-3-8b-v1_1-hf| model~\footnote{\url{https://huggingface.co/xtuner/llava-llama-3-8b-v1_1-transformers}}, which is fine-tuned from the LLaMA-3-8B architecture. This variant of LLaVA was optimized using the AdamW~\cite{ADAMW} optimizer, with a learning rate set to $1e-5$ and a batch size of 4.
    \item \textbf{MiniCPM-V.} For the task of figure description extraction, we used the latest model (\texttt{MiniCPM-Llama3-V 2.5}~\footnote{\url{https://huggingface.co/openbmb/MiniCPM-Llama3-V-2_5}}) in the MiniCPM-V families without the fine-tuning.
    \item \textbf{LLaMA-3-8B.} To generate a caption, we used \verb|Meta-Llama-3-8B-Instruct|~\footnote{\url{https://huggingface.co/meta-llama/Meta-Llama-3-8B-Instruct}} model. The model was trained with AdamW, a learning rate of $1e-5$ and a batch size of 1.
    \item \textbf{Yi-1.5-9B} is an upgraded version of Yi. For generating captions, we used \verb|Yi-1.5-9B-Chat|~\footnote{\url{https://huggingface.co/01-ai/Yi-1.5-9B-Chat}}. The training configuration is the same as LLaMa-3-8B.
    \item \textbf{Pegasus.} We trained the \verb|pegasus-large|~\footnote{\url{https://huggingface.co/google/pegasus-large}} model using the AdamW optimizer, with a batch size of 32 and a learning rate of $5e-5$.
\end{itemize}

When fine-tuning the caption generation models, we concatenated all figure-mentioning paragraphs. In cases where the cumulative length of these paragraphs exceeded 512 tokens, the text was truncated to fit within this limit. For text generation, we utilized greedy search to ensure the generation of coherent.

\begin{table}[t]
\caption{Comparison of generated captions. MLBCAP includes key information and provides comprehensive descriptions, whereas Pegasus tends to produce shorter captions that lack sufficient detail.}
\renewcommand{\arraystretch}{1.5}
\begin{adjustbox}{width=\textwidth,center}
\begin{tabular}{cm{3.5cm}m{8cm}} \hline
\textbf{Figure} & \multicolumn{1}{c}{\textbf{Model}} & \multicolumn{1}{c}{\textbf{Caption}} \\ \hline 
\multirow{5}{*}{\raisebox{-5.0cm}{\includegraphics[width=5.5cm]{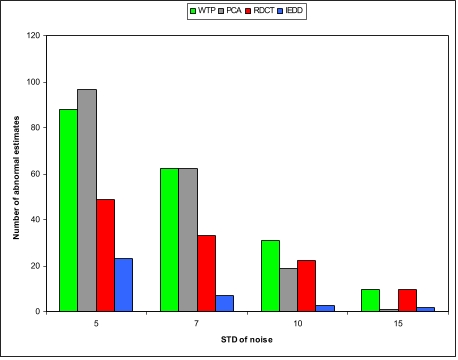}}} 
 & \centering \small{\textbf{MLBCAP (long)}}  & Fig. 4. IEDD significantly outperforms other methods in reducing abnormal estimates. The bar chart shows that IEDD consistently has the lowest number of abnormal estimates across various noise levels (STD of noise: 5, 7, 10, 15), compared to WTP, PCA, and RDCT. \\ \cline{2-3} 
 & \centering \small{\textbf{MLBCAP (short)}} & Fig. 4. IEDD significantly outperforms other methods in reducing abnormal estimates across various noise levels (STD of noise: 5, 7, 10, 15). \\ \cline{2-3}
 & \centering \small{LLaMA-3-8B} & Fig. 4. Abnormal estimates (AE) for different methods. The proposed method IEDD significantly outperforms all compared methods. The results are obtained for the synthetic dataset. \\ \cline{2-3}
 & \centering \small{Yi-1.5-9B} & Fig. 4. Number of abnormal estimates (AE) for different noise variances. For each variance, the experiment was repeated 100 times. The proposed method IEDD significantly outperforms other compared methods. \\ \cline{2-3}
 & \centering Pegasus & Fig. 4. Comparison of RRMSE of the proposed method (IEDD) with other methods. \\ \hline
\end{tabular}
\end{adjustbox}
\label{tab:mlbcap_example}
\end{table}

\subsection{Human Evaluation Results}
We evaluate MLBCAP with a human evaluation to accurately assess the quality of generated captions.
The baseline models used for evaluation include LLaMA-3-8B, Yi-1.5-9B, and Pegasus. All baselines were fine-tuned on the high-quality captions from the SciCap and SciCap+ datasets, specifically captions with a quality score of 5 or higher. The qualitative results for generated captions are illustrated in Table~\ref{tab:mlbcap_example}.

\subsection{Comparison with Baselines} 
We recruited three computer vision (CV) experts, each with over five years of experience in the field. These experts were asked to select the best and worst quality captions generated by LLMs. We used 91 CV figures from the SciCap test dataset, with the candidate captions de-identified and randomly shuffled for each figure before being presented to the experts.

\begin{figure}[t]
\centering

\begin{minipage}[t]{0.45\textwidth}
    \centering
    \includegraphics[width=\textwidth]{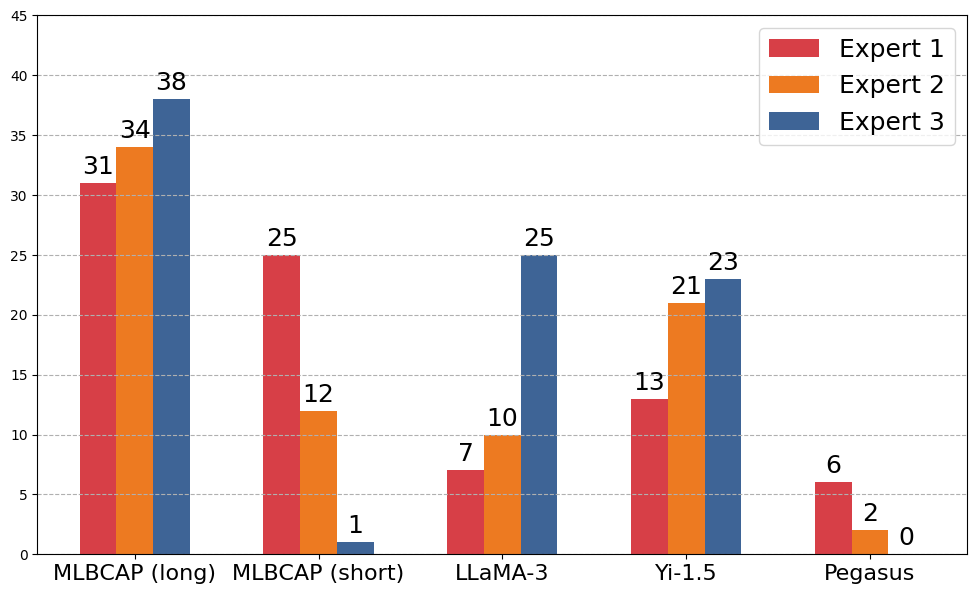}
    \textbf{(a)} Frequency of high-quality captions. 
    \label{fig:human_high}
\end{minipage}
\hfill 
\begin{minipage}[t]{0.45\textwidth}
    \centering
    \includegraphics[width=\textwidth]{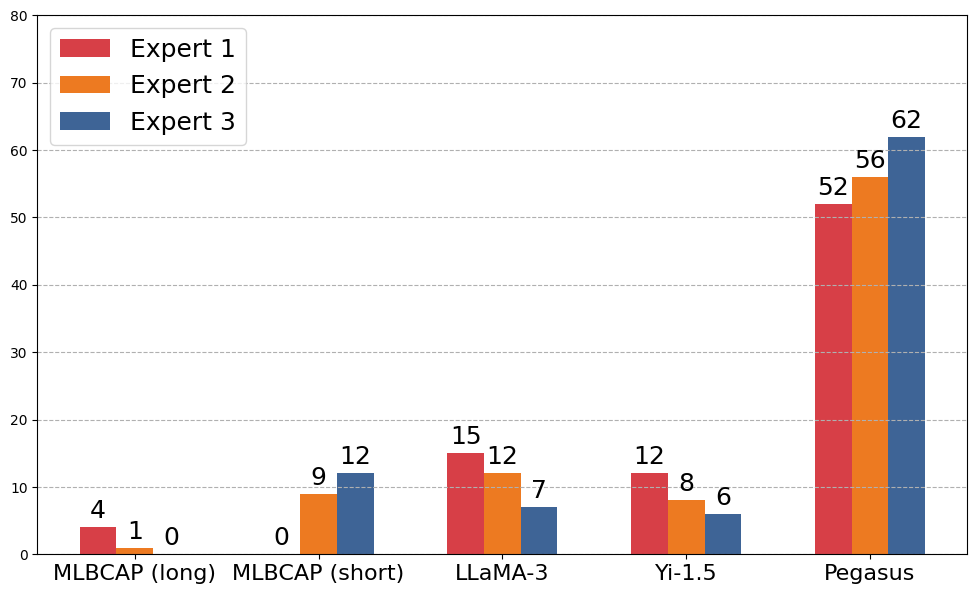}
    \textbf{(b)} Frequency of low-quality captions. 
    \label{fig:human_low}
\end{minipage}

\caption{The human evaluation results for selecting captions based on quality. Captions generated by MLBCAP (long) are most frequently selected as high quality.}
\label{fig:human_eval}
\end{figure}

Figure~\ref{fig:human_eval}(a) presents the results of the human evaluation. Notably, captions generated by MLBCAP (long) were consistently selected as high-quality by all three experts, indicating a strong preference compared to baselines. To understand the high preference for the MLBCAP, we investigated which LLM-generated captions were selected during the best caption selection phase. The percentages of captions selected from the GPT-4o, Yi-1.5-9B, LLaMA-3-8B, and Pegasus models were 89.38\%, 4.23\%, 6.17\%, and 0.19\%, respectively. As expected, GPT-4o performed exceptionally well in generating high-quality captions.

Interestingly, the assessment of MLBCAP (short) captions revealed variability in expert opinion.
Despite this variance, Figure~\ref{fig:human_eval}(b) demonstrates that MLBCAP (short) captions were rarely selected as the worst quality, suggesting that the differences in preference may stem more from individual biases towards caption length rather than from a significant discrepancy in caption quality.

\subsection{Comparison with Author Captions} 
Another human evaluation was conducted as part of the 2nd SciCap challenge~\footnote{\url{http://scicap.ai/}}.
To account for the natural distinctions and ensure equitable assessment between short and long captions, the task was divided into two tracks. Participants in the short caption track were required to submit results where at least 30\% of the captions were no longer than the original author-written captions from the SciCap test dataset. Similarly, teams in the long caption track were required to submit results where at least 30\% of the captions exceeded the length of the original author-written captions. In the case of MLBCAP, 68.15\% of the captions in the long caption track were longer than the author-written captions, while 46.53\% of the captions in the short caption track were shorter than the author's captions.

Three judges (who are not the CV experts), all native American English speakers with expertise in technical academic writing, were recruited to rank the captions.
They ranked captions for the same 200 figures, randomly selected from the challenge test set, based on how effectively they convey the figure’s message for each track. The selection of these figures adhered to the following criteria: (1) For the long caption track, figures were selected where the author-written captions were shorter than the generated captions submitted by participants. (2) For the short caption track, figures were chosen where the author-written captions exceeded the length of the generated captions submitted by participants. (3) For both the long and short caption tracks, only figures with a SciCap-Eval score of 4 or higher were included.

As shown in Table~\ref{tab:challenge}, in the long caption track, MLBCAP (long) outperformed all others, receiving the best score of 1.27. This result presents the effectiveness of our approach in generating comprehensive captions that resonated well with expert evaluators. Additionally, although the MLBCAP's caption in the short caption track ranked marginally lower than the author-written captions, it still placed our method ahead of other methods.

\begin{table}[t]
\centering
\caption{The human evaluation results for the SciCap challenge and MLBCAP is compared with the second-best solutions for each track. Lower average ranks indicate higher preference by the judges.}
\begin{adjustbox}{width=8.5cm,center}
\renewcommand{\arraystretch}{1.3}
\begin{tabular}{lcc} \hline
\textbf{Team} & \textbf{Long Caption} & \textbf{Short Caption} \\ \hline
Author-written & 2.84 & \textbf{1.52} \\ \hline
LM-Ensemble~\cite{LMENSEMBLE} & 2.82 & 3.56 \\
Length-Adaptive LLM~\cite{LengthAdaptiveLLM} & 3.08 & 3.18 \\ \hline
MLBCAP & \textbf{1.27} & 1.74 \\ \hline
\end{tabular}
\end{adjustbox}
\label{tab:challenge}
\end{table}

\subsection{Analysis} Here we conduct an analysis focused on the components of MLBCAP to examine their impact on caption quality, using a random sample of 200 instances from the SciCap test dataset.
Table~\ref{tab:ablation_01} illustrates the significant improvements in caption quality across all models when figure descriptions and a robust filtering process are incorporated. While models such as LLaMA-3-8B and Yi-1.5-9B perform well in generating high-quality captions, GPT-4o emerges as the best model for providing top-tier candidate captions.

However, Table~\ref{tab:ablation_02} reveals a crucial insight; the highest caption quality score (5.440) is achieved not solely by relying on the individual model (GPT-4o) but through a strategic combination of multiple LLMs and subsequent post-editing. This approach demonstrates that even with the availability of a powerful model like GPT-4o, the incorporation of diverse perspectives from smaller LLMs can yield improved results.

\begin{table}[t]
\centering
\caption{The impact of including figure descriptions and the filtering process. Each value is the caption quality score (SciCap-Eval).}
\begin{adjustbox}{width=8.5cm,center}
\renewcommand{\arraystretch}{1.3}
\begin{tabular}{lccc} \hline
  & \textbf{LLaMA-3-8B} & \textbf{YI-1.5-9B} & \textbf{GPT-4o} \\ \hline
Base model & 4.612 & 4.575 & 5.305 \\
+ Quality Assessment & 4.905 & 4.910 & - \\
+ Figure description & \textbf{5.030} & \textbf{5.005} & \textbf{5.390} \\ \hline
\end{tabular}
\end{adjustbox}
\label{tab:ablation_01}
\end{table}

\begin{table}[t]
\caption{Ablation study analyzing the effect of the best caption selection and post-editing on caption quality.}
\begin{adjustbox}{width=6cm,center}
\renewcommand{\arraystretch}{1.3}
\begin{tabular}{cccc} \hline
\textbf{Multi-LLM} & \textbf{Post-edit} & \textbf{\begin{tabular}[c]{@{}c@{}}Quality Score\\ (SciCap-Eval)\end{tabular}} \\ \hline
\ding{56} (GPT-4o) & \ding{56} & 5.390  \\ \hline
\ding{56} (GPT-4o) & \ding{52} & 5.405  \\ \hline
\ding{52} & \ding{56} & 5.430  \\ \hline
\ding{52} & \ding{52} & \textbf{5.440}  \\ \hline
\end{tabular}
\end{adjustbox}
\label{tab:ablation_02}
\end{table}

\section{Discussion}

\subsection{Caption Preferences}

While MLBCAP rarely produced low-quality captions in human evaluations, we observed variability among experts when selecting the best and worst captions. This challenge was reflected in the low inter-rater agreement metrics. For instance, Fleiss' kappa for selecting high-quality captions among baseline models was 0.154, indicating low agreement, while for low-quality captions, the kappa improved to 0.382, signifying moderate agreement. Similarly, in the SciCap Challenge, Kendall's tau for inter-rater agreement, provided by the challenge organizers, was 0.3589 for long captions and 0.1100 for short captions, highlighting the difficulty of reaching a consensus even among experienced judges.

These findings align with previous research~\cite{SciCapSum}, which has shown that caption ranking tasks inherently elicit varied judgments from evaluators. The low agreement underscores the complexity of defining ``quality'' in figure captioning, where multiple factors interplay, such as the caption’s informativeness, length, detail, and overall style.

A plausible explanation for the observed discrepancies lies in the subjective nature of caption preferences. Evaluators may prioritize different attributes, such as the level of detail versus brevity, or favor stylistic differences in language. For instance, one expert might value a caption that provides exhaustive detail, while another might prefer concise summaries that align with the space constraints typical in scientific publications. This subjectivity in preferences naturally leads to variability in judgments, reducing inter-rater reliability.

These observations highlight the importance of developing clearer guidelines and evaluation criteria for figure captions. A more standardized framework could help align evaluators' judgments and establish a consensus on what constitutes a ``high-quality'' caption. Future work could explore leveraging LLMs not only for caption generation but also as assistive tools for evaluating captions in a more consistent manner, thereby addressing some of the subjectivity inherent in human evaluations. This would ensure that assessments of caption quality are both rigorous and aligned with the needs of diverse scientific communities.

\begin{table}[t]
\caption{Evaluation results for the SciCap test dataset. The ROUGE (F1-score) and BLEU (4-gram) scores of MLBCAP are opposite to the human preference (Figure~\ref{fig:human_eval}).}
\begin{adjustbox}{width=9cm,center}
\renewcommand{\arraystretch}{1.3}
\begin{tabular}{ccccc} \hline
        & \textbf{ROUGE-1}  & \textbf{ROUGE-2} & \textbf{ROUGE-L} & \textbf{BLEU-4} \\ \hline
Pegasus   & \textbf{0.460} & \textbf{0.282} & \textbf{0.418} & 0.124         \\
LLaMA-3-8B     &   0.405     &   0.237         &      0.346    & 0.126     \\
Yi-1.5-9B     &    0.412    &   0.244         &     0.354      & \textbf{0.134}    \\ \hline
MLBCAP (short)     &   0.369     &   0.174         &    0.310  & 0.043         \\
MLBCAP (long) &   0.333     &   0.150         &    0.257   & 0.049        \\ \hline       
\end{tabular}
\end{adjustbox}
\label{tab:performance_rouge}
\end{table}

\subsection{Evaluation with Traditional Metrics} 

Furthermore, we found a discrepancy between traditional metric-based evaluations and human judgments. In Table~\ref{tab:performance_rouge}, while MLBCAP was preferred over the author-written captions in the human evaluations, traditional metrics failed to capture the perceived quality of captions. This divergence presents the limitations of relying solely on conventional metrics for evaluating caption quality. Similar observations have been made in the natural image captioning task~\cite{chan-etal-2023-clair}, where BLEU and ROUGE scores showed low correlation with human judgments (Kendall's tau of approximately 0.3 for both metrics).

This inconsistency can be attributed to the inherent limitations of BLEU and ROUGE metrics, which focus on n-gram overlap between the generated caption and the reference caption. In scientific figure captioning, there are multiple valid ways to describe the same content using different terminologies or phrasings. As a result, even high-quality captions may receive low scores if they do not closely match the reference.

\section{Limitations}
While our experiments indicate the potential of MLBCAP in the figure captioning task, there are some limitations that point to possible directions for future work.

Firstly, the integration of multiple LLMs in our framework introduces a trade-off between performance and efficiency. The reliance on multiple models not only reduces inference speed but also increases the demand for computational resources. This may limit the practical scalability of MLBCAP, particularly in environments with restricted computational capabilities or in real-time applications.

Secondly, MLBCAP incorporates a closed-source LLM as a critical component of the caption generation pipeline. This inclusion imposes inherent limitations, particularly in terms of transparency and interpretability. The closed-source nature restricts our ability to fully understand and analyze the model’s reasoning processes and decision-making behavior, which may hinder trust and adoption in certain scientific communities where explainability is crucial.

Lastly, our evaluation was primarily conducted through human assessments on arXiv papers, which, while valuable, does not fully capture the generalization capabilities of MLBCAP across a broader range of scientific literature. To rigorously validate the robustness and adaptability of the model, future evaluations should include a diverse set of scientific documents.

\section{Conclusion} 

In this paper, we presented MLBCAP, a novel framework for generating high-quality captions for scientific figures through the collaborative utilization of multiple Large Language Models (LLMs). Unlike prior approaches that rely on isolated modalities or limited data perspectives, MLBCAP uniquely integrates textual and visual features alongside a filtering mechanism to ensure that only high-quality training data is utilized. By combining the complementary strengths of multiple LLMs with candidate caption generation and a post-editing stage, our framework generates captions that are not only preferred over author-written captions in informativeness but also cater to diverse needs through long and short caption formats. In addition, our results highlight the effectiveness of a multi-LLM approach, demonstrating higher caption quality compared to a single prominent LLM like GPT-4o.

\section{Acknowledgement}

We are grateful to the anonymous reviewers for their valuable feedback. This research, along with the SciCap Challenge 2024, was partially supported by the Alfred P. Sloan Foundation (Grant Number: 2024-22721).

\bibliographystyle{splncs04}
\bibliography{ref}

\appendix

\newpage
\section*{Appendix}

\begin{table}[]
\caption{The actual prompts we used and the text in magenta is a placeholder. In part, \texttt{[Figure]} is an image that is used as an input for MLLMs.}
\centering
\begin{tabularx}{\linewidth}{>{\hsize=0.3\hsize\centering\arraybackslash}X
                               >{\hsize=0.7\hsize\arraybackslash}X}
\hline
\textbf{Purpose}  &  \textbf{Prompt}  \\ \hline

Quality Assessment  
&  \small{\textcolor{magenta}{\texttt{[Figure]}}

\#\#\# Paragraphs

\textcolor{magenta}{\texttt{[Paragraphs]}}

\#\#\# Caption

\textcolor{magenta}{\texttt{[Caption]}}

Given the figure, paragraphs and caption, please rate the level of usefulness of the caption from 1 to 6 based on how well the caption could help readers understand the important information. 6 is the highest. 1 is the lowest. The answer should be JSON format: \{``rating'': \}.}
\\ \midrule
\vspace{3pt}

\begin{tabular}[c]{@{}c@{}}Figure Description\\ (GPT-4o)\end{tabular} 
& \small{\textcolor{magenta}{\texttt{[Figure]}}

Your task is to describe a figure from a scientific paper.

Answer your results in JSON format.

\#\#\# Background

Figure is a \textcolor{magenta}{\texttt{[Figure Type]}}.

It is a figure about the topic \textcolor{magenta}{\texttt{[Subject]}}.

\#\#\# Rule

Description of the figure should be accurate and clear. If in doubt, avoid numerical expressions. Provide the description in JSON format with the following key: description.}
\\ \hline

\begin{tabular}[c]{@{}c@{}}Figure Description\\ (MiniCPM-V)\end{tabular}
& \small{\textcolor{magenta}{\texttt{[Figure]}} 

What is in the image?}
\\   \hline
\end{tabularx}
\label{tab:actual_prompts_01}
\end{table}

\begin{table}[]
\caption{The prompt for the caption generation. In part, \texttt{[Figure]} is an image that is used as an input for MLLMs.}
\centering
\begin{tabularx}{\linewidth}{>{\hsize=0.3\hsize\centering\arraybackslash}X
                               >{\hsize=0.7\hsize\arraybackslash}X}
\hline
\textbf{Purpose}  &  \textbf{Prompt}  \\ \hline

\begin{tabular}[c]{@{}c@{}}Caption Generation\\ (Few-shot)\end{tabular}
& \small{Your task is to create a caption that summarizes based on a paragraph.

\#\#\# Figure Caption

The format of a Figure Caption is Declarative title + Description + Statistical
information (optional).

Declarative title: summarises the result or major finding of the data you are presenting in the
figure. (A mere representation of the x and y axes cannot be a title.)

Description: a brief description of the results necessary for understanding the figure without
having to refer to the main text

Statistical information: for example, number of replicates, asterisks denoting P-values,
statistical tests, etc.

\#\#\# Background

Figure is a \textcolor{magenta}{\texttt{[Figure Type]}}.

Figure is a category related to \textcolor{magenta}{\texttt{[Subject]}}.

\#\#\# Rule

Caption MUST have a word count of 60 words or less.

Caption MUST have a tone and sentence structure appropriate for a top-tier conference (e.g.,
NeurIPS, ICLR, CVPR, ACL, EMNLP).

It is not a caption to describe the x-axis y-axis.

Caption MUST be clear, concise, consistent, and provide specific information, especially not
false.

If the given paragraph uses abbreviations, use them in the caption.

\#\#\# Best Caption Examples

\textcolor{magenta}{[Few-shot Examples]}

\#\#\# Input

Paragraph:
\textcolor{magenta}{[Paragraphs]}

Figure Summary:
\textcolor{magenta}{[Figure Description]}

Mention:
\textcolor{magenta}{[Mentions]}

OCR text:
\textcolor{magenta}{[OCR]}

\#\#\# Output format

Answer results in JSON format: \{``caption'': \}.
}
\\ \hline
\end{tabularx}
\label{tab:actual_prompts_01}
\end{table}

\begin{table}[]
\caption{The prompt for the selecting best caption and post-editing process. The \texttt{[Max Len]} is the constraint of caption lengths (Long: 50, Short: 30).}
\centering
\begin{tabularx}{\linewidth}{>{\hsize=0.15\hsize\centering\arraybackslash}X
                               >{\hsize=0.85\hsize\arraybackslash}X}
\hline
\textbf{Purpose}  &  \textbf{Prompt}  \\ \hline

Judgement  
&  \small{
A good figure caption should include the following elements:

1. **Clear Description**: Clearly describe what the figure represents so that readers can understand
the main point of the figure just by reading the caption.

2. **Conciseness**: Keep it concise while including all essential information. The caption MUST be
brief yet informative (important!!).

3. **Relevant Information**: Include background information, experimental conditions, or methods
used that are necessary to understand the figure. This helps the reader interpret the data correctly.

4. **Consistency**: Maintain consistency with the rest of the paper in terms of terminology and style.
Ensure that the terms used in the caption match those used in the text.

5. **Citation**: If necessary, include citations of related research or references in the paragraph.

You are given a summarization of the figure, relevant paragraphs, a mentioned sentences, and four
caption candidates:

\#\#\# Summarization of the Figure

\textcolor{magenta}{[Figure Description]}

\#\#\# Paragraph

\textcolor{magenta}{[Paragraphs]}

\#\#\# Mention

\textcolor{magenta}{[Mentions]}

\#\#\# Caption A

\textcolor{magenta}{[Pegasus Caption]}

\#\#\# Caption B

\textcolor{magenta}{[LLaMA-3-8B Caption]}

\#\#\# Caption C

\textcolor{magenta}{[Yi-1.5-9B Caption]}

\#\#\# Caption D

\textcolor{magenta}{[GPT-4o Caption]}

1. Choose the best and worst caption and answer in JSON format (For example, if A is the best and B
is the worst, the answer is: {{"Good": "A", "Bad": "B}}). Candidate captions shouldn't be scored low
just because they're concise.

2. If even the best caption could be improved, use the candidate captions and paragraphs to improve it
(math symbols, legend, grammar, etc.).

3. The improved sentence should have a tone and sentence structure appropriate for a top-tier
conference (e.g., NeurIPS, ICLR, CVPR, ACL, EMNLP) and MUST have a word count of \textcolor{magenta}{[Max Len]} words or less.

4. If you find that sentences are becoming long and complex, making it difficult for readers to
understand, break the sentences up to effectively convey the important information.

5. If you already provided a perfect caption, keep it the same.

6. Do not omit the figure numbers, such as in "Fig. 3" or "Figure 5".
Provide them in JSON format with the following keys: Good, Bad,
Improved Caption
{{"Good" : "", "Bad" : "", "Improved Caption": ""}}

}
\\ \hline
\end{tabularx}
\label{tab:actual_prompts_02}
\end{table}

\end{document}